\documentclass[letterpaper,twocolumn]{jpsj3}
\usepackage{txfonts}
\usepackage{url}
\usepackage{xcolor}

\usepackage{latexsym}
\usepackage{amsmath, amssymb}
\usepackage{braket}
\usepackage{cite}
\usepackage{bm}
\usepackage[ruled,vlined]{algorithm2e}
\usepackage{algpseudocode}
\usepackage{enumitem}
\usepackage{booktabs}
\usepackage{tabularx}
\usepackage{array}
\usepackage{cleveref}
\usepackage{graphicx}

\title{Improving FMQA via Initial Training Data Design Considering Marginal Bit Coverage in One-Hot Encoding}

\author{Taiga Hayashi$^{1}$\thanks{taiga.hayashi@keio.jp}, Yuya Seki$^{1,2}$, Kotaro Terada$^{3}$, Yosuke Mukasa$^{1,3}$, Shuta Kikuchi$^{1,2}$, and Shu Tanaka$^{1,2,4,5}$\thanks{shu.tanaka@keio.jp}}
\inst{
$^{1}$Graduate School of Science and Technology, Keio University, Yokohama, Kanagawa 223-8522, Japan \\
$^{2}$Keio Sustainable Quantum Artificial Intelligence Center, Keio University, Minato-ku, Tokyo 108-8345, Japan \\
$^{3}$Quanmatic Inc., Shinjuku-ku, Tokyo 162-0042, Japan\\
$^{4}$Department of Applied Physics and Physico-Informatics, Keio University, Yokohama, Kanagawa 223-8522, Japan \\
$^{5}$Human Biology-Microbiome-Quantum Research Center, Keio University, Minato-ku, Tokyo 108-8345, Japan \\
}

\abst{
Factorization machine with quadratic-optimization annealing (FMQA) is a black-box optimization method that combines a factorization machine (FM) surrogate with QUBO-based search by an Ising machine. 
When FMQA is applied to integer or discretized continuous variables via one-hot encoding, uniform random initial sampling can leave many binary variables never active in the initial training data, and the corresponding FM parameters receive no direct gradient updates from the observed responses. 
We address this by designing the initial training data to achieve complete marginal bit coverage, namely, ensuring that every binary variable obtained by one-hot encoding takes the value one at least once. 
We use two space-filling sampling methods, Latin hypercube sampling (LHS) and the Sobol' sequence, yielding LHS-FMQA and Sobol'-FMQA. 
On the human-powered aircraft wing-shape optimization benchmark with 17 and 32 design variables, both proposed methods achieved numerically higher mean final cruising speeds than the baseline FMQA, with the advantage more pronounced on the 32-variable problem.
}

\begin{document}
    \maketitle
    \section{Introduction}
\label{Sec:Introduction}

A combinatorial optimization problem is the problem of finding a combination of discrete decision variables that minimizes or maximizes an objective function under various constraints. Such problems appear in many areas of society, including delivery scheduling in logistics, network configuration in communication systems, taxi matching, financial portfolio optimization, materials design, and shift scheduling in manufacturing~\cite{Dantzig1959TruckDispatching,Laporte2009FiftyYearsVRP,MagnantiWong1984NetworkDesign,AlonsoMora2017RideSharing,markowitz2008portfolio,Rosenberg2016TradingQuantumAnnealer,Burke2004NurseRostering,Lookman2019ActiveLearningMaterials}. As the number of decision variables increases, the number of candidate solutions grows exponentially, and exhaustive search becomes extremely difficult for large-scale problems. Therefore, a practical search method is required to avoid combinatorial explosion while still obtaining good solutions.

To address this demand, Ising machines, which implement metaheuristic algorithms for combinatorial optimization on hardware, have been extensively studied~\cite{Mohseni2022IsingMachinesReview,Tanahashi2019IsingMachines}. To solve a combinatorial optimization problem using an Ising machine, the target problem is formulated as an Ising model or a mathematically equivalent quadratic unconstrained binary optimization (QUBO) model~\cite{Lucas2014,Chakrabarti2017Quantum,Inoue2021,BorosHammer2002PseudoBoolean,Anthony2017QuadraticReformulations,Fred2022qubo}. This formulation allows the Ising machine to perform solution search as an energy-minimization problem while avoiding exhaustive enumeration of all candidate solutions. Many Ising machines introduce stochastic fluctuations during search and gradually reduce them to promote convergence to low-energy states. Thus, Ising machines can perform efficient approximate search even for large-scale problems while mitigating the increase in computational time caused by combinatorial explosion.

On the other hand, some combinatorial optimization problems have complex constraints or objective functions, making direct formulation as an Ising model or QUBO difficult or practically impossible. A function for which the analytical relationship between inputs and outputs is unknown, and for which only the output $f_{\mathrm{BB}}(\bm{x})$ for a given input $\bm{x}$ can be observed, is called a black-box (BB) function. The problem of finding an input $\bm{x}^{\ast}$ that minimizes or maximizes the objective function value using only observed input-output pairs is called black-box optimization (BBO). In experimental and simulation-based BBO problems that require sequential trial and error, each evaluation of the BB function may involve substantial time or financial cost. Therefore, it is important to obtain better solutions with a limited number of function evaluations. When a target value is specified, reducing the number of function evaluations required to reach that target is also important. As the number of decision variables increases, the search space rapidly expands, and the number of evaluations required to reach a high-quality solution tends to increase.

BBO requires an efficient search framework for obtaining good solutions under a limited evaluation budget. Representative approaches to BBO include genetic algorithms (GAs), which are stochastic search algorithms~\cite{Holland1975Adaptation,Goldberg1989GA,BackFogelMichalewicz1997HandbookEC}, and Bayesian optimization (BO), which is a surrogate-model-based method~\cite{Shahriari2016BOReview}. These methods have been widely studied and have played important roles in many applications.

As an approach for efficiently solving BBO problems, methods that use Ising machines even when the objective function itself cannot be directly formulated as a QUBO have attracted attention. Factorization machine with quadratic-optimization annealing (FMQA)~\cite{Kitai2020,tamura2026blackbox} has been proposed as a discrete BBO method that employs a factorization machine (FM)~\cite{Rendle2010} as a surrogate model and performs optimization using an Ising machine or another quadratic optimization solver. Because the FM output can be converted into a QUBO form, FMQA enables search by an Ising machine even when the BB function cannot be directly converted into a QUBO formulation.

FMQA has been applied to various domains, including materials design~\cite{Kitai2020,Nawa2023MTJ,Kim2024OpticalDiodeFMQA}, engineering design~\cite{Matsumori2022QUBO,Inoue2022PCSEL}, network optimization~\cite{furusawa2026comparative}, and feature selection~\cite{Tamura2024Polypropylene,Kikuchi2026Epistasis}. Previous studies have reported that FMQA can obtain high-quality solutions with fewer evaluations than random search, GAs, particle swarm optimization, and BO~\cite{Kitai2020,tamura2026blackbox,kikuchi2026rnafmqa}. These reports suggest that FMQA is a promising framework when each BB function evaluation is expensive and only a limited number of evaluations are available.

Several studies have investigated factors that affect the optimization performance of FMQA and related methods. These studies can be roughly classified into three categories: encoding design, surrogate-model regularization, and training-data usage. In the first category, for integer-variable BBO problems, the influence of integer-to-binary encoding methods has been studied, and one-hot encoding has been reported to be effective for small-scale problems~\cite{seki2022blackboxoptimizationintegervariableproblems}. The influence of binary-integer encoding and variable assignment has also been investigated in an FMQA framework for RNA inverse folding~\cite{kikuchi2026rnafmqa}. In addition, an appropriate mapping from problem variables to binary variables can affect the search landscape and improve solution quality~\cite{Koshikawa2025BitLabeling}. In the second category, for continuous-variable problems, function smoothing regularization has been proposed to improve the precision of FMQA~\cite{FSRFMQA.7.013149}. In the third category, methods that modify how the accumulated training dataset is used have also been proposed, including SWIFT-FMQA, which uses a sliding-window strategy to retain recently added data points~\cite{nakano2026}, and subsampling factorization machine annealing, which trains the FM using sampled subdatasets~\cite{hama2025subsamplingfactorizationmachineannealing}. These studies clarify important aspects of FMQA, but the design of the initial training data itself has received much less attention.

This gap is particularly important in FMQA because binary encoding is not merely a data-preprocessing choice but a component that directly determines both FM training inputs and the QUBO searched by the Ising machine. 
In FMQA for integer or discretized continuous variables, the search variables must be encoded in binary form before being passed to the Ising machine. 
Among the possible integer-to-binary encodings, one-hot encoding has been reported to perform well in FMQA~\cite{seki2022blackboxoptimizationintegervariableproblems} and is commonly used. 
Under one-hot encoding, exactly one bit takes the value one for each original variable, while all other bits take the value zero.
Therefore, if the initial training data are generated by uniform random sampling without further consideration, some bits may never take the value one in the initial dataset. Since the gradient of the FM output with respect to parameters associated with a bit is proportional to the value of that bit, parameters corresponding to bits that remain zero do not receive direct dataset-derived updates~\cite{FSRFMQA.7.013149}. As a result, the estimated QUBO coefficients for those bits can depend strongly on the initial parameter values, which may introduce bias into the subsequent search by the Ising machine. In this sense, the unresolved problem addressed in this paper is not which encoding should be adopted, but how the initial training data should be designed once one-hot encoding is used in FMQA.

The objective of this study is to improve the optimization performance of FMQA by enhancing marginal bit coverage in the initial training data, that is, ensuring that every binary variable obtained by one-hot encoding takes the value one at least once. Since FMQA trains an FM surrogate model from the initial training data and then searches for candidate solutions on the approximated model, the initial dataset can strongly affect both search efficiency and final solution quality. To realize complete marginal bit coverage, we introduce initial data generation methods based on Latin hypercube sampling (LHS) and the Sobol' sequence, and evaluate them on a human-powered aircraft (HPA) wing-shape optimization benchmark. Here, LHS and Sobol' sequences are used as means to construct such initial datasets, rather than as ends in themselves.

The remainder of this paper is organized as follows. Section~\ref{Sec:FMA} reviews FMQA, including the model equation of the FM, parameter update, and the optimization flow. Section~\ref{Sec:proposed_method} presents the proposed initial training data generation method based on quasi-random sequences and explains how it ensures complete marginal bit coverage in one-hot encoding. Section~\ref{Sec:Method} describes the human-powered aircraft wing-shape optimization benchmark, integer-to-binary conversion, and the experimental settings used in this study. Section~\ref{Sec:result} shows the numerical results for the benchmark problem. Section~\ref{Sec:Discussion} discusses the relationship between marginal bit coverage in the initial training data and the optimization performance of FMQA. Finally, Sec.~\ref{Sec:Conclusion} concludes the paper and describes future perspectives.

    \section{FMQA}
\label{Sec:FMA}

FMQA is a BBO method that combines an FM, which is a machine-learning model, with an Ising machine. In FMQA, an FM is trained on observed input-output data to approximate the BB function. Next, an Ising machine searches for a combination of decision variables that minimizes the output of the trained FM, and the obtained candidate input is evaluated by the BB function and added to the training data. By iterating these steps, FMQA aims to improve the best-found objective function value under a limited evaluation budget.

\subsection{Model Equation of the FM}
\label{Subsec:FM_equation}

In FMQA, an FM is used as a surrogate model of the BB function. Because an FM can be expressed as a quadratic polynomial in binary variables, it is mathematically equivalent to a QUBO model. This property makes it possible to input the minimization problem of the trained FM output into an Ising machine and perform efficient solution search. Therefore, the FM is a machine-learning model that has a high affinity with BBO algorithms that use Ising machines.

The model equation of the FM is given by
\begin{equation}
  \label{eq:FM}
  \begin{aligned}
    f_{\mathrm{FM}}(\bm{x};\bm{\theta})
    &= \omega_0
    + \sum_{i=1}^N \omega_i x_i
    + \sum_{1\le i < j \le N}\langle \bm v_i,\,\bm v_j \rangle x_i x_j.
  \end{aligned}
\end{equation}
Here, $\bm{x}\equiv(x_1,\dotsc,x_N)\in\{0,1\}^N$ is a set of $N$ binary variables $x_i$, and $\bm{\theta}\equiv(\omega_0,\{\omega_i\}_{i=1}^{N},\{\bm v_i\}_{i=1}^{N})$ is the set of model parameters. Also, $\omega_0\in\mathbb{R}$, $\omega_i\in\mathbb{R}$, and $\bm{v}_i\in\mathbb{R}^{K}\ (i=1,\dotsc,N)$ are parameters updated through training, and $K$ is a hyperparameter. 
The FM is used as a surrogate for the BB function. Since the FM takes a binary vector ${\bm x} \in \{0,1\}^N$ as input while the original BB function $f_{\rm BB}$ is defined on the continuous (or integer) decision variable space, the FM parameters are trained so that
\begin{align}
    f_{\rm FM}({\bm x}_d;{\bm \theta}) \approx f_{\rm BB}(\phi({\bm x}_d)),
\end{align}
holds for each training input ${\bm x}_d$, where $\phi:\mathcal{X}_{\rm OH} \to \mathbb{R}^{n_x}$ is the decoding map from one-hot binary vectors to the corresponding decision variables. Here, $n_x$ denotes the dimension of the design variables, and the binary input is structured as $N = n_x M$ bits, with each of the $n_x$ design variables represented by $M$ binary variables. The set of valid one-hot configurations is
\begin{align}
    \mathcal{X}_{\rm OH} = \left\{ \bm{x} \in \{0,1\}^{n_x M} \,\middle|\,
    \sum_{m=0}^{M-1} x_{j,m} = 1, \ \forall j = 1, \ldots, n_x \right\}.
\end{align}
The explicit form of $\phi$ is given in Sec.~\ref{subsec:integer-to-binary-conversion}.
The inner product $\langle \bm{v}_i,\,\bm{v}_j \rangle$ in $\mathbb{R}^{K}$ is defined as
\begin{equation}
  \langle \bm{v}_i,\,\bm{v}_j \rangle
  = \sum_{k=1}^{K} v_{i,k}\,v_{j,k},
\end{equation}
where $v_{i,k}$ denotes the $k$-th component of the vector $\bm{v}_i$.
FM is a machine-learning model that can represent feature interactions in a low-rank form for sparse data with many zero-valued bits and can effectively approximate nonlinear dependencies that conventional linear models cannot capture.
The model equation of the FM can be reduced to a QUBO form. Specifically, using the FM parameters $\omega_i$ and $\bm{v}_i$, the QUBO matrix $Q$ is defined as
\begin{equation}
\label{eq:Qij}
  Q_{i,j} =
  \begin{cases}
    \omega_i,                               & (i = j), \\[4pt]
    \langle \bm {v}_i,\,\bm {v}_j \rangle,  & (i \ne j).
  \end{cases}
\end{equation}
Since adding a constant to the objective does not change the location of its minimum, the bias term $\omega_0$ can be ignored when minimizing $f_{\rm FM}$ via QUBO. We therefore consider only the linear and quadratic terms.

Then, the QUBO energy $E_{\mathrm{QUBO}}(\bm{x})$ is written as
\begin{equation}\label{eq:E_qubo}
  \begin{aligned}
    E_{\mathrm{QUBO}}(\bm{x})
      &= \sum_{1 \le i \le j \le N} Q_{i,j}\,x_i\,x_j \\[4pt]
      &= \sum_{i=1}^N Q_{i,i}\,x_i^2
       + \sum_{1 \le i < j \le N} Q_{i,j}\,x_i\,x_j \\[4pt]
      &= \sum_{i=1}^N \omega_i\,x_i
       + \sum_{1 \le i < j \le N} \langle \bm {v}_i,\,\bm {v}_j\rangle\,x_i\,x_j.
  \end{aligned}
\end{equation}
Here, we used $x_i^2=x_i$ because $\bm{x}\in\{0,1\}^N$. Therefore, the part of $f_{\mathrm{FM}}(\bm{x};\bm{\theta})$ excluding the constant term $\omega_0$ coincides with $E_{\mathrm{QUBO}}(\bm{x})$, and minimizing the FM output can be treated as QUBO minimization by an Ising machine.

\subsection{FM Parameter Update}
\label{Subsec:FM_parameter_update}

This subsection describes FM parameter updates using the mean squared error (MSE) as the loss function and estimates the computational cost required for the updates. In the numerical experiments of this study, AdamW~\cite{Kingma2015Adam,Loshchilov2019AdamW} is used as the FM parameter optimization method.

Let $\{\bm{x}_d\}_{d=1}^{D} \subset \mathcal{X}_{\rm OH}$ denote a set of $D$ training inputs.
The training dataset $\mathcal{D}$ is then defined as
\begin{equation}
  \mathcal{D}=\{(\bm{x}_d,\,f_{\mathrm{BB}}(\phi(\bm{x}_{d})))\}_{d=1}^{D},
\end{equation}
where $D$ denotes the number of training data points accumulated up to the current FMQA iteration.
FM training uses the MSE between the FM prediction value $f_{\mathrm{FM}}(\bm{x}_d;\bm{\theta})$ and the output $f_{\mathrm{BB}}(\phi(\bm{x}_{d}))$ of the BB function as the loss function.
\begin{align}
\label{eq:FM_loss}
L(\bm{\theta})
&=
\frac{1}{D}\sum_{d=1}^D
\Bigl(
  f_{\mathrm{FM}}(\bm{x}_d;\bm{\theta})\;-\;f_{\mathrm{BB}}(\phi(\bm{x}_{d}))
\Bigr)^2.
\end{align}
AdamW is used to update the FM parameters. Let the estimated parameter vector be $\hat{\bm{\theta}}$ and the learning rate be $\eta>0$. The update rule is given by
\begin{align}
\label{eq:FM_update}
\bm{m}_t &= \beta_1 \bm{m}_{t-1} + (1-\beta_1)\bm{g}_t ,\\
\bm{v}_t &= \beta_2 \bm{v}_{t-1} + (1-\beta_2)\bm{g}_t \odot \bm{g}_t ,\\
\hat{\bm{m}}_t &= \frac{\bm{m}_t}{1-\beta_1^t}, \\
\hat{\bm{v}}_t &= \frac{\bm{v}_t}{1-\beta_2^t} ,\\
\hat{\bm{\theta}}_{t+1}
&=
\hat{\bm{\theta}}_t
- \eta \frac{\hat{\bm{m}}_t}{\sqrt{\hat{\bm{v}}_t}+\epsilon}
- \eta \lambda \hat{\bm{\theta}}_t .
\label{eq:FM_update_theta}
\end{align}
Here, $\bm{g}_t$ denotes the gradient of $L$ with respect to $\bm{\theta}$ at iteration $t$; $\bm{m}_t$ and $\bm{v}_t$ are exponentially moving averages of $\bm{g}_t$ and $\bm{g}_t \odot \bm{g}_t$, with decay rates $\beta_1$ and $\beta_2$, respectively; $\hat{\bm{m}}_t$ and $\hat{\bm{v}}_t$ are their bias-corrected versions; $\epsilon$ is a small constant for numerical stability; $\lambda$ is the weight-decay coefficient; and $\odot$ denotes the element-wise product.
By iterating the update rule in Eqs.~\eqref{eq:FM_update}--\eqref{eq:FM_update_theta}, we estimate $\hat{\bm{\theta}}$ that reduces the error between the FM output $f_{\mathrm{FM}}(\bm{x}_d;\hat{\bm{\theta}})$ and the observed value $f_{\mathrm{BB}}(\phi(\bm{x}_d))$ for the training data.

Next, we confirm under what conditions each parameter contributes to the update. The gradient of the loss function in Eq.~\eqref{eq:FM_loss} is
\begin{align}
\label{eq:grad_loss}
\dfrac{\partial L(\bm{\theta})}{\partial \bm{\theta}}
=
\dfrac{2}{D}\sum_{d=1}^D
\Bigl(
  f_{\mathrm{FM}}(\bm{x}_d;\bm{\theta})-f_{\mathrm{BB}}(\phi(\bm{x}_{d}))
\Bigr)
\dfrac{\partial f_{\mathrm{FM}}(\bm{x}_d;\bm{\theta})}{\partial \bm{\theta}},
\end{align}
and the gradient of the loss function is determined by the partial derivatives of the FM output with respect to each parameter. From the FM model equation in Eq.~\eqref{eq:FM}, these partial derivatives are given by
\begin{equation}
\label{eq:partial_F}
  \frac{\partial f_{\mathrm{FM}}(\bm{x};\bm{\theta})}{\partial \theta} =
  \begin{cases}
    1 & (\theta = \omega_0), \\[2pt]
    x_i & (\theta = \omega_i),\\[2pt]
    \Bigl(\sum_{j=1}^N v_{j,l}x_j\Bigr)x_i - v_{i,l}x_i^2
    & (\theta = v_{i,l}).
  \end{cases}
\end{equation}
Here, $l=1,\dotsc,K$. From Eq.~\eqref{eq:partial_F}, the partial derivative with respect to $\omega_i$ is proportional to $x_i$, and the partial derivative with respect to $v_{i,l}$ also contains $x_i$. Therefore, when $x_i=0$ at a certain data point, that data point does not provide a direct gradient to the parameters corresponding to bit $i$. This property means that, in one-hot encoding, which bits take the value one directly affects the training of the FM parameters.

Finally, we estimate the computational cost of gradient calculation. In the third expression of Eq.~\eqref{eq:partial_F}, the term $\sum_{j=1}^N v_{j,l}x_j$ is independent of $i$ and can therefore be computed once for each $l$ and shared. Thus, computing $\sum_{j=1}^N v_{j,l}x_j$ requires $\mathcal{O}(KN)$ operations, and computing the partial derivatives for all $i$ and $l$ after that also requires $\mathcal{O}(KN)$ operations. Therefore, the computational cost of one gradient calculation per data sample is $\mathcal{O}(KN)$, so FM updates remain efficient even though pairwise feature interactions are taken into account.

We next specify the FM parameter initialization used throughout this study. 
The interaction matrix parameters $\{v_i\}$ are initialized by independent standard normal random variables, the linear coefficients $\{\omega_i\}$ are initialized by Xavier uniform initialization~\cite{Glorot2010Xavier}, and the bias parameter $\omega_0$ is initialized to zero.

A consequence of using AdamW with decoupled weight decay is that even parameters that receive no data-driven gradient updates may evolve through the weight-decay term in Eq.~\eqref{eq:FM_update_theta}. 
Such parameters gradually drift toward zero rather than remaining at their initial values. 
We therefore distinguish between parameters that are mathematically frozen and parameters that, although they may change during training, are not informed by the observed BB responses. 
This distinction is used in Sec.~\ref{Subsec:variety_of_bits}.

\subsection{Optimization Flow of FMQA}
\label{Subsec:Flow_of_FMQA}

The BBO procedure using FMQA is summarized as follows.

\begin{description}[leftmargin=0pt,labelindent=0pt,labelwidth=0pt,labelsep=0pt,style=nextline]
  \item[\textbf{Step 1: Generation of initial training data}]
  Initial inputs $\{\bm{x}_{d}\}_{d=1}^{N_0}$ for the BB function are generated and evaluated to obtain the initial dataset
  \begin{equation}
    \mathcal{D}_0=\{(\bm{x}_{d},f_{\mathrm{BB}}(\phi(\bm{x}_{d})))\}_{d=1}^{N_0}.
  \end{equation}

  \item[\textbf{Step 2: Training of the FM surrogate model}]
  Using the dataset $\mathcal{D}_n$, the FM model $f_{\mathrm{FM}}(\bm{x};\hat{\bm{\theta}}_n)$ is trained to approximate the BB function value $f_{\mathrm{BB}}(\phi(\bm{x}))$. Here, $\hat{\bm{\theta}}_n$ denotes the estimated parameter values updated so that the MSE becomes small.

  \item[\textbf{Step 3: Candidate search using an Ising machine}]
  Because the trained FM model $f_{\mathrm{FM}}(\bm{x};\hat{\bm{\theta}}_n)$ is a quadratic form, it can be converted into the QUBO matrix $Q$ using Eq.~\eqref{eq:Qij}. An Ising machine is then used to search for a candidate solution $\bm{x}_{\mathrm{new}}$ that minimizes $E_{\mathrm{QUBO}}(\bm{x})$.

  \item[\textbf{Step 4: Evaluation of the BB function and dataset update}]
  The candidate $\bm{x}_{\mathrm{new}}$ is decoded by $\phi$ and evaluated by the BB function to obtain $f_{\mathrm{BB}}(\phi(\bm{x}_{\mathrm{new}}))$. This data point is added to the dataset as
  \begin{equation}
    \mathcal{D}_{n+1}=\mathcal{D}_n\cup\{(\bm{x}_{\mathrm{new}},\,f_{\mathrm{BB}}(\phi(\bm{x}_{\mathrm{new}})))\}.
  \end{equation}
  The procedure then returns to Step 2.
\end{description}

By repeating these iterations until either a predetermined upper limit on the number of evaluations is reached or a target value is achieved, FMQA aims to obtain high-quality solutions.

    \section{Proposed Method}
\label{Sec:proposed_method}
The objective of this study is to suppress the degradation of search performance caused by bias in the training data, and to thereby improve its optimization performance. To this end, we propose an extended FMQA framework whose initial training data are generated so that every binary variable obtained by one-hot encoding takes the value one at least once, and we then run FMQA on this dataset.

\subsection{Marginal Bit Coverage in the Initial Training Dataset}
\label{Subsec:variety_of_bits}

We begin by formalizing the notion of bit coverage that motivates our proposed initialization. For an initial training dataset $\mathcal{D}_0 = \{\bm{x}_d\}_{d=1}^{N_0}$ with one-hot encoded inputs $\bm{x}_d \in \{0,1\}^N$, let $c_{j,m} = \sum_{d=1}^{N_0} x_{d,j,m}$ denote the number of times that the $m$-th binary variable obtained by one-hot encoding of design variable $j$ takes the value one in the initial dataset. The proposed initialization methods guarantee complete marginal bit coverage, that is, $c_{j,m} \ge 1$ for all $j = 1, \ldots, n_x$ and $m = 0, \ldots, M-1$.

From Eqs.~\eqref{eq:FM_loss}--\eqref{eq:partial_F}, FM training uses the MSE between the objective function values and the FM predictions as the loss function and updates the model parameters by minimizing it. Since the input $\bm{x}$ consists of binary variables discretized by integer-to-binary conversion, some FM parameters corresponding to bits that take the value zero at a data point do not receive direct dataset-derived gradients from that data point~\cite{FSRFMQA.7.013149}. Therefore, when one-hot encoding is used, any bias regarding which bits take the value one in the dataset directly affects the training process.

In general, let the number of design variables be $n_x$, the number of discrete values for each variable be $M$, and the number of initial training data points be $N_0$. When each design variable is represented by $M$ binary variables using one-hot encoding and $N_0$ initial training data points are generated uniformly at random, the probability that an arbitrary binary variable never takes the value one in the initial training data is given by
\begin{equation}
\label{eq:prob_untrained_bit}
\left(1-\frac{1}{M}\right)^{N_0}.
\end{equation}
Therefore, the expected number of binary variables that never take the value one in the initial training data is
\[
n_x M \left(1-\frac{1}{M}\right)^{N_0}.
\]
This quantity provides a quantitative measure of how much bias can remain in the initial training data.
The numerical implication of this estimate for the HPA benchmark setting used in this study is given in Sec.~\ref{Sec:Method}.

Now consider the situation where a bit $i$ never takes the value one across the $D$ training data points. That is, assume that
\begin{equation}
x_{d,i}=0\quad(\forall d = 1, \dotsc , D)
\end{equation}
holds for the training data $\bm{x}_d$. Here, $x_{d,i}$ is the $i$-th component of $\bm{x}_d$. In this case, from Eq.~\eqref{eq:partial_F}, the gradients of the parameters corresponding to this bit are
\begin{align}
 &\frac{\partial f_{\mathrm{FM}}(\bm{x}_{d};\bm{\theta})}{\partial \omega_i}=0\ \quad(\forall d), \\
 &\frac{\partial f_{\mathrm{FM}}(\bm{x}_{d};\bm{\theta})}{\partial v_{i,\ell}}=0\ \quad(\forall d,\ \forall \ell).
\end{align}
Substituting these into Eq.~\eqref{eq:grad_loss}, the data-driven gradients of the loss function with respect to the corresponding FM parameters also vanish:
\begin{equation}
\frac{\partial L(\bm{\theta})}{\partial \omega_i}=0,\qquad
\frac{\partial L(\bm{\theta})}{\partial v_{i,\ell}}=0\quad(\forall \ell).
\end{equation}
However, this does not imply that $\omega_i$ and $v_{i,\ell}$ remain constant during training: under AdamW, the decoupled weight-decay term $-\eta \lambda \hat{\bm{\theta}}_t$ in Eq.~\eqref{eq:FM_update_theta} updates these parameters even in the absence of data-driven gradients, gradually pulling them toward zero. 
We therefore refer to such parameters as parameters that are not informed by the observed BB responses, distinguishing them from parameters that are mathematically frozen.
As a result, the QUBO coefficients corresponding to such FM parameters are not informed by the observed BB responses; instead, they are determined only by the initialization and the subsequent weight-decay dynamics.

Thus, if bias occurs in the bits that take the value one in the dataset, the corresponding FM parameters are not informed by the observed BB responses. 
The associated QUBO coefficients in Eq.~\eqref{eq:Qij} therefore carry little information about the BB function in the missing-bit directions, and the candidate search by the Ising machine over those directions is effectively uninformed. 
As a result, the quality of candidate solution selection may decrease, leading to degradation of optimization performance.

\subsection{Space-Filling Initial Designs for Marginal Bit Coverage}
\label{Subsec:quasi_random}

We adopt two space-filling sampling methods to construct initial training data with complete marginal bit coverage: LHS~\cite{McKay1979LHS,Bingham2015LHS}, a stratified sampling method with randomization, and the Sobol' sequence~\cite{Sobol1967}, a deterministic low-discrepancy sequence. We describe how each method, when applied with $N_0 = M$, achieves complete marginal bit coverage under one-hot encoding.
Rather than detailing the full generation procedures, we focus on the conditions under which each method, combined with one-hot encoding, ensures that every binary variable obtained by one-hot encoding takes the value one at least once.
Let $\bm{s}_t=(s_{t,1},\ldots,s_{t,n_x})$ denote the normalized continuous point generated by LHS or the Sobol' sequence for the $t$-th initial sample.
This continuous point is not directly used as the design variable value evaluated by the BB function.
Instead, each coordinate is first converted into a discrete index
\begin{equation}
q_{t,j}=\min\{\lfloor M s_{t,j}\rfloor,\ M-1\}
\quad (q_{t,j}\in\{0,\ldots,M-1\}),
\end{equation}
and the one-hot variable used as the FMQA input is then defined as
\begin{equation}
x_{t,j,m}=\mathbf{1}[q_{t,j}=m].
\end{equation}
The actual design variable value evaluated by the BB function is the discrete value $z_{q_{t,j}}^{(j)}$ defined in Sec.~\ref{subsec:integer-to-binary-conversion}.

LHS is a stratified sampling method that divides the domain of each variable into as many strata as the number of samples and extracts one point from each stratum~\cite{McKay1979LHS,Bingham2015LHS}. 
Therefore, when (i) $M$ discrete values are assigned to each variable in one-hot encoding, (ii) the LHS strata are aligned one-to-one with these discrete values through the above map $s_{t,j}\mapsto q_{t,j}\mapsto x_{t,j,m}$, and (iii) $N_0=M$ is chosen, LHS samples exactly one point from each stratum.
Consequently, each discrete value is selected at least once, and each corresponding binary variable obtained by one-hot encoding takes the value one at least once. Thus, LHS with $N_0=M$ is effective for constructing initial training data in which all binary variables obtained by one-hot encoding take the value one at least once.
Note, however, that while LHS guarantees that each marginal interval of each design variable is sampled exactly once, the random pairing of intervals across different design variables means that pairwise combinations of intervals are sampled only sparsely. 

The Sobol' sequence is a deterministic low-discrepancy sequence whose first $2^p$ points form a $(0, p, 1)$-net in base 2 in each
one-dimensional projection~\cite{Sobol1967,niederreiter1992}, i.e., when $[0, 1)$ is partitioned into $2^p$ equal subintervals, each subinterval contains exactly one of the first $2^p$ points.
Consider the setting $N_0 = M = 2^p$, and let
\begin{equation}
  s_t = (s_{t,1}, \ldots, s_{t,n_x}) \in [0, 1)^{n_x}
  \quad (t = 1, \ldots, N_0),
\end{equation}
denote the first $N_0$ Sobol' points. With $I_m = [m/M, (m+1)/M)$, the definition above gives $x_{t,j,m}=1$ if and only if $s_{t,j}\in I_m$. Therefore, the $(0, p, 1)$-net property guarantees that for each $(j, m)$, exactly one of the $N_0$ points satisfies $x_{t,j,m}=1$.
Therefore, every binary variable obtained by one-hot encoding takes the value one exactly once in the initial training data, achieving complete marginal bit coverage.
In the actual implementation for statistical evaluation, we use scrambled Sobol' points with different random seeds to obtain different initial datasets across runs. Here, scrambling means a randomized bitwise transformation of the binary digits of Sobol' points that preserves their hierarchical base-2 structure~\cite{owen1995randomly}, so that the low-discrepancy character of the Sobol'-based initialization is retained.

These constructions guarantee only marginal bit coverage and do not guarantee coverage of pairwise bit combinations. The FM model in Eq.~\eqref{eq:FM} contains pairwise interaction terms $\langle v_i, v_j \rangle x_i x_j$, and the corresponding coupling is informed by data only when bits $i$ and $j$ are simultaneously one. Under a valid one-hot representation, two bits belonging to the same design variable cannot be simultaneously one. Therefore, the number of potentially co-active bit pairs is not $\binom{n_x M}{2}$ but
\begin{equation}
\binom{n_x M}{2} - n_x\binom{M}{2}
=
\binom{n_x}{2}M^2.
\end{equation}
Because each initial data point activates only one bit per design variable, the initial dataset can cover only a limited subset of these cross-variable bit pairs. The proposed initialization therefore ensures that the FM parameters associated with individual bits are exposed to active data points, but it does not ensure that all pairwise bit co-occurrences are observed. We return to the implications of this limitation in Sec.~\ref{Sec:Discussion}.

The above guarantee of complete marginal bit coverage requires $N_0 = M$ (and additionally $M = 2^p$ for the Sobol' construction).
Note that complete marginal coverage requires $N_0 \ge M$ under valid one-hot encoding. 
When the evaluation budget for initial sampling is smaller than $M$, partial coverage strategies (e.g., prioritizing the most influential variables) may be considered, but complete coverage is unattainable.
We leave such extensions to future work. 
In this study, we focus on the practically common case $N_0 = M$.

    \section{Experimental Setup}
\label{Sec:Method}

This section describes the method used in this study. First, we outline the HPA benchmark problem, which is treated as a BB function in this study. Next, we describe the integer-to-binary conversion.

\subsection{HPA Benchmark}
\label{subsec:HPA}

In this study, we adopt the HPA benchmark problem~\cite{Namura2025HPA} as the BB function for FMQA. HPA is known as an engineering optimization problem for the main-wing design of a human-powered aircraft, and the objective function is evaluated by aeroelastic analysis that considers the interaction between aerodynamics and structural mechanics. In this coupled analysis, the aerodynamic load acting on the wing causes structural deformation, and the deformation in turn affects the aerodynamic characteristics. Because this bidirectional relationship is treated simultaneously, the numerical computation cost is high and the evaluation is complex. Furthermore, the objective function in HPA is not given as an analytical form, i.e., neither the functional shape nor gradient information is accessible. In addition, the optimal solution is unknown. 

Therefore, HPA serves as a fully BB function for which both the optimal solution and the function landscape are unknown. It thus provides a realistic and challenging benchmark for evaluating sequential surrogate-model-based BBO methods, and we adopt it in this study.

The HPA benchmark includes three single-objective optimization problems: HPA101, whose objective function is the required power; HPA102, whose objective function is the wing weight; and HPA103, whose objective function is the cruising speed. Each of HPA101, HPA102, and HPA103 has three types: level 1 with 17 design variables, level 2 with 32 design variables, and level 3 with 108 design variables. The suffixes ``-1'', ``-2'', and ``-3'' appended to the problem names correspond to these levels. In this study, among these, we target HPA103-1, which has 17 design variables, and HPA103-2, which has 32 design variables. As reported by Namura~\cite{Namura2025HPA}, HPA103 exhibits stronger multimodality and higher optimization difficulty than HPA101 and HPA102, and HPA103-2 contains many local optima or deceptive structures. Namura~\cite{Namura2025HPA} also noted that problems of this type may pose challenges for surrogate-model approximation. Since approximating the BB function by an FM is expected to be particularly difficult for such problems, we consider HPA103-1 and HPA103-2 to be appropriate testbeds for evaluating the optimization performance of FMQA.
The HPA103 problem is originally formulated as a maximization of the cruising speed $v(\bm{z})$, where $\bm{z} \in \mathbb{R}^{n_x}$ denotes the vector of $n_x$ continuous design variables.
In this study, we treat it as a minimization problem by defining the BB function as $f_{\rm BB}(\bm{z}) = -v(\bm{z})$. 
Throughout the paper, all reported objective values in tables and figures correspond to $v(\bm{z})$ in m/s, even though the optimization internally minimizes $f_{\rm BB}$.
The HPA benchmark implementation used in this study is the publicly available code accompanying Namura~\cite{Namura2025HPA}. 
The design variable bounds for HPA103-1 (17 variables) and HPA103-2 (32 variables) follow the specifications in~\cite{Namura2025HPA}. 
The aeroelastic evaluation is deterministic, so any variability across trials in our experiments arises solely from the initial training data and the stochasticity of the Ising-machine sampler. 
Designs that violate the implicit physical constraints of the simulator are penalized by the benchmark itself, following the convention in~\cite{Namura2025HPA}.

\subsection{Integer-to-Binary Conversion}
\label{subsec:integer-to-binary-conversion}

The HPA problem targeted in this study is expressed by continuous variables, but FMQA uses an Ising machine for solution search, so the optimization variables must be given as discrete variables. In other words, continuous variables cannot be handled directly, and it is necessary to convert them into integer and binary variables in advance. Previous studies have reported that, among integer-to-binary conversion methods in FMQA, one-hot encoding shows high optimization performance~\cite{seki2022blackboxoptimizationintegervariableproblems}. 
Therefore, in this study, after discretizing each continuous variable to an integer variable, we adopt one-hot encoding to convert the integer variables into binary variables suitable for solution search by an Ising machine.

Before one-hot encoding, $M$ equally spaced discrete values are set on the domain $[z_{\min}^{(j)},z_{\max}^{(j)}]$ of each continuous variable $z^{(j)}$, and the discretized integer variable is defined as
\begin{equation}
  q_j \in \{0,1,\dotsc,M-1\}.
\end{equation}
Here, it is assumed that $q_j=m$ corresponds to the $m$-th discrete value of the $j$-th design variable, given by
\begin{equation}
\label{eq:zm-discretization}
    z_m^{(j)} = z_{\rm min}^{(j)}
    + \frac{m}{M-1}
      \left( z_{\rm max}^{(j)} - z_{\rm min}^{(j)} \right).
\end{equation}
That is, $m=0$ corresponds to $z_{\min}$ and $m=M-1$ corresponds to $z_{\max}$. In one-hot encoding, for the $j$-th design variable, $M$ binary variables
\begin{equation}
  x_{j,0},\dots,x_{j,M-1}\in\{0,1\}
\end{equation}
are introduced to represent this $q$. If
\begin{equation}
\label{eq:onehot-constraint}
  \sum_{m=0}^{M-1} x_{j,m} = 1
\end{equation}
holds, the index $m$ for which $x_{j,m}=1$ is used to represent $q_j=m$. That is, exactly one component takes the value one and the others take the value zero.
Combining Eqs.~(\ref{eq:zm-discretization}) and~(\ref{eq:onehot-constraint}), the decoding map $\phi$ introduced in Sec.~\ref{Subsec:FM_equation} is explicitly given, for $\bm{x} \in \mathcal{X}_{\rm OH}$, by
\begin{align}
    [\phi(\bm{x})]_j
    = z_{\rm min}^{(j)}
    + \frac{m_j(\bm{x})}{M - 1}
      \left( z_{\rm max}^{(j)} - z_{\rm min}^{(j)} \right),
    \label{eq:phi-explicit}
\end{align}
where $m_j(\bm{x}) \in \{0, 1, \ldots, M-1\}$ is the unique index satisfying $x_{j, m_j(\bm{x})} = 1$, and $[z_{\rm min}^{(j)}, z_{\rm max}^{(j)}]$ denotes the domain of the $j$-th design variable.
The total number of binary variables in the FM input is therefore $N = n_x M$. For HPA103-1 with $n_x = 17$ and $M = 32$, this yields $N = 544$; for HPA103-2 with $n_x = 32$, it yields $N = 1024$. The FMQA search is performed on a discretized grid of the original continuous design space, with grid resolution determined by $M$.

In the experimental setting used in this study, $M=32$ and $N_0=32$. Therefore, if the initial training data are generated uniformly at random, the probability that a given binary variable obtained by one-hot encoding never takes the value one is
\[
\left(1-\frac{1}{32}\right)^{32}
=
\left(\frac{31}{32}\right)^{32}
\approx 0.364.
\]
Thus, about 36.4\% of the binary variables are expected to be absent from the initial training data at the marginal level. In terms of the total number of binary variables, this corresponds to about 198 out of $17\times 32=544$ variables for HPA103-1 and about 372 out of $32\times 32=1024$ variables for HPA103-2.

\subsection{Settings of the FM and Ising Machine}

Table~\ref{tab:hpa_fm_ising} shows the settings of the FM and Ising machine for the optimization of HPA103 with FMQA. Here, $n_x$ denotes the dimension of the design variables, that is, the total number of continuous design variables in HPA103. 
The objective function $f_{\rm BB}$ used in FMQA is defined in Sec.~\ref{subsec:HPA}.
At each FMQA iteration, the FM is initialized anew and trained from scratch using all training data accumulated up to that iteration; the parameter values from the previous iteration are not used as a warm start.
The objective given to the Ising machine is the FM output augmented with a penalty for the one-hot constraints:
\begin{equation}
E_{\mathrm{IM}}(\bm{x})
=
f_{\mathrm{FM}}(\bm{x};\hat{\bm{\theta}})
+
\lambda_{\mathrm{pen}}
\sum_{j=1}^{n_x}
\left(\sum_{m=0}^{M-1} x_{j,m} - 1\right)^2.
\end{equation}
Also, in this study, we use the penalty method that adds constraint conditions such as the one-hot constraint to the objective function, and its weight is denoted by the penalty coefficient $\lambda_{\mathrm{pen}}$, which is defined as
\begin{equation}
\lambda_{\mathrm{pen}} = 8\cdot \max~\!\bigl(1,\ \lfloor \max |f_{\mathrm{BB}}(\phi(\bm{x}))|+0.5 \rfloor\bigr).
\end{equation}
Here, $\lambda_{\mathrm{pen}}$ denotes the penalty coefficient for the one-hot constraints. Also, $\max |f_{\mathrm{BB}}(\phi(\bm{x}))|$ denotes the maximum absolute value of the objective function over all training data accumulated up to the current iteration, not only over the initial training data.
We also use Fixstars Amplify AE~\cite{FixstarsAmplify2025}, a simulated annealing sampler implemented on a GPU, as the Ising machine. The solver timeout is set to 2000\,ms, and the other parameters are left at their default values.
Because the one-hot constraints are imposed through a penalty term rather than as hard constraints, the sampler may in principle return a binary assignment that does not strictly satisfy them. In the implementation used in this study, such an assignment is decoded blockwise into an integer candidate by selecting the active bit in each block; if multiple bits are active, the first active bit is used, and if no bit is active, one index is selected uniformly at random in that block. In the present experiments, however, no such one-hot-constraint violation was observed in the sampled solutions.
We also avoid reevaluating candidates that have already been included in the training data.
If the solution returned by the Ising machine is identical to an evaluated binary vector, it is first decoded into integer indices, each index is perturbed by a random value in $\{-1,0,1\}$ and clipped to the valid range, and the resulting integer vector is encoded again into a one-hot binary vector.
This replacement is repeated until an unevaluated candidate is obtained, and only this candidate is evaluated by the BB function and added to the training data.
\begin{table}[t]
\centering
\footnotesize
\caption{Settings of the FM and Ising machine}
\label{tab:hpa_fm_ising}
\renewcommand{\arraystretch}{1.0}
\setlength{\tabcolsep}{6pt}
\begin{tabularx}{\linewidth}{@{} p{0.35\linewidth} p{0.20\linewidth} X @{}}
\toprule
Item & Parameter & Value \\
\midrule
Dimension & $n_x$ & $17$ (-1), $32$ (-2) \\
Number of discrete values & $M$ & 32 \\
Number of initial samples & $N_0$ & 32 \\
Evaluations & -- & 200 \\
Hyperparameter & $K$ & 5 \\
Optimizer & -- & AdamW \\
Learning rate & $lr$ & 0.5 \\
AdamW coefficients & $\beta_1$, $\beta_2$, $\epsilon$, $\lambda$ & $0.9$, $0.999$, $10^{-8}$, $0.01$ \\
Batch size & -- & 8 \\
Number of epochs & -- & 500 \\
Penalty coefficient & $\lambda_{\mathrm{pen}}$ & $8\cdot \max~\!\bigl(1,\ \lfloor \max |f_{\mathrm{BB}}(\phi(\bm{x}))|+0.5 \rfloor\bigr)$ \\
Timeout & -- & $2000\,\mathrm{ms}$ \\
\bottomrule
\end{tabularx}
\end{table}

In the proposed method, the initial training data are generated by LHS~\cite{McKay1979LHS,Bingham2015LHS} and the Sobol' sequence~\cite{Sobol1967}. The detailed generation method is described in Sec.~\ref{Sec:proposed_method}. To guarantee that every binary variable obtained by one-hot encoding takes the value one at least once in the initial training data, we set the number of initial samples equal to the number of discrete values, i.e., $N_0=M$. For the Sobol' sequence, this guarantee additionally requires $M=N_0=2^p$, because it relies on the $(0,p,1)$-net property of the first $2^p$ points in each one-dimensional projection.

\subsection{Comparison Methods}

In this study, to evaluate the optimization performance of the proposed method, we compare it with an FMQA baseline that does not enforce marginal bit coverage in the initial training data, denoted Conv-FMQA. In this baseline, the initial training data are generated by uniform random sampling.
As a comparison with classical optimization algorithms, we adopt GP-BO, a Bayesian optimization method, and NSGA-II, a genetic algorithm. 
NSGA-II is originally a multi-objective evolutionary algorithm. We include it here as a representative GA-based optimizer in single-objective mode, given its standard implementation in Optuna~\cite{10.1145/3292500.3330701}. 
For both GP-BO and NSGA-II, we use the standard implementations provided in Optuna~\cite{10.1145/3292500.3330701}.
In addition, Random Search is included as a simple baseline.

For GP-BO, NSGA-II, and Random Search, the number of initial data points is also set to $N_0=32$, the same as in FMQA. Their initial data are generated by uniform random sampling in the original continuous design space. We do not use the LHS- or Sobol'-based initializations for these classical optimizers, because those initializations are introduced here specifically to control marginal bit coverage after one-hot encoding in FMQA.
Random Search is implemented in the same continuous-variable formulation.
For all three methods, the termination criterion is based on the total number of BB function evaluations, and the evaluation budget is set to 200 so as to match the FMQA setting. In addition, the design variables are optimized directly in the original continuous design space, and optimizer-specific hyperparameters other than the common evaluation budget and the number of initial data points are left at their default values.
We note an important asymmetry between FMQA-based methods and the comparison methods: FMQA-based methods perform optimization on the discretized $M$-level grid imposed by one-hot encoding, whereas GP-BO, NSGA-II, and Random Search operate on the original continuous design space. 
The primary comparison in this study is therefore between Conv-FMQA and the proposed FMQA variants (LHS-FMQA, Sobol'-FMQA), which share the same discretized search space and FMQA framework. GP-BO, NSGA-II, and Random Search are included as reference baselines representing established BBO frameworks under continuous-variable formulations.
The comparison with these methods should be interpreted with this asymmetry in mind.

    \section{Results}
\label{Sec:result}

\begin{figure}[t]
  \centering
  \includegraphics[width=\linewidth]{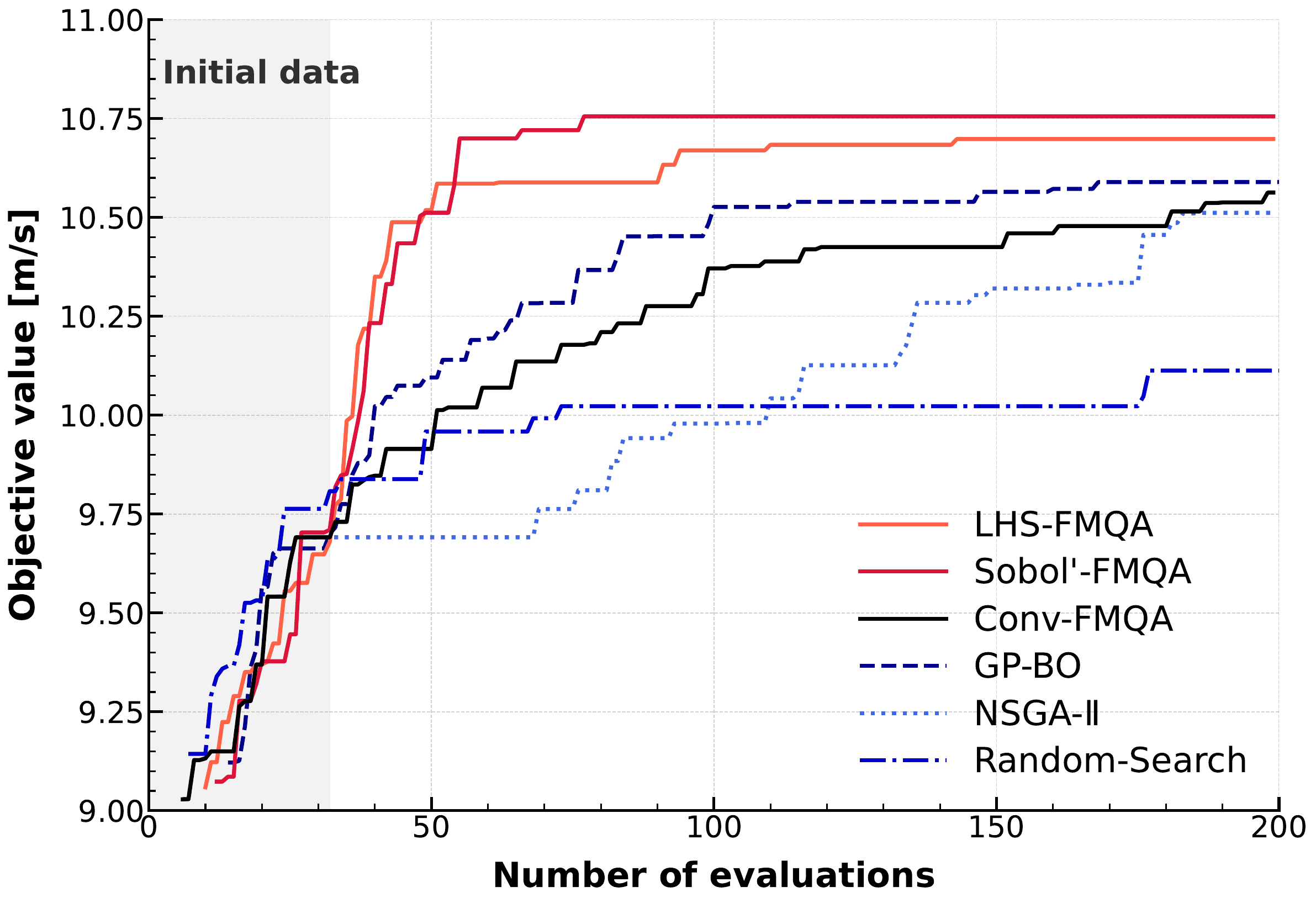}
  \caption{Comparison of the optimization processes of each method for HPA103-1. The horizontal axis represents the number of function evaluations, and the vertical axis represents the best value found. The gray shaded region corresponds to the evaluations used to construct the initial training dataset.}
  \label{fig:HPA1031_opt_result}
\end{figure}

\begin{figure}[t]
  \centering
  \includegraphics[width=\linewidth]{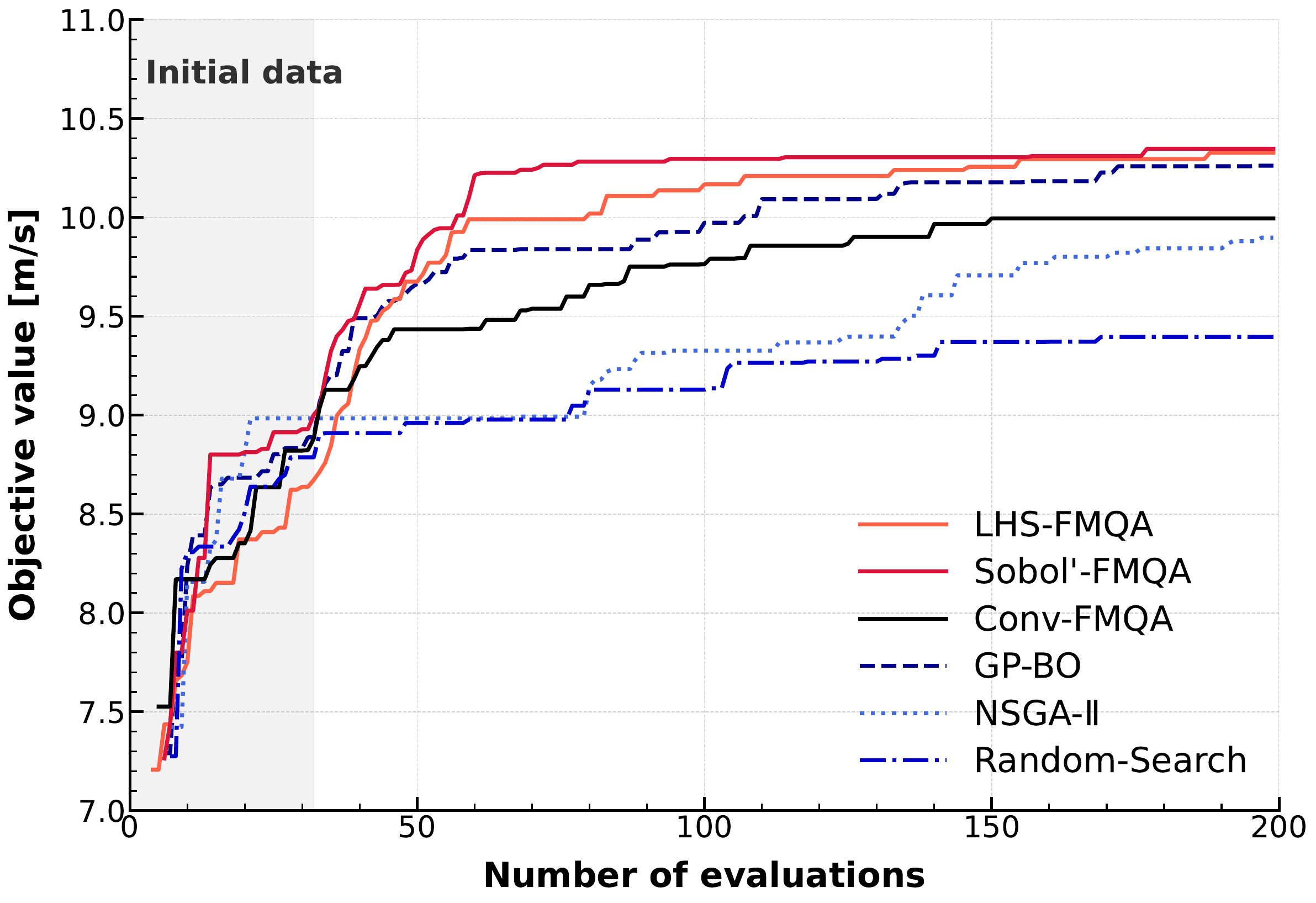}
  \caption{Comparison of the optimization processes of each method for HPA103-2. The horizontal axis represents the number of function evaluations, and the vertical axis represents the best value found. The gray shaded region corresponds to the evaluations used to construct the initial training dataset.}
  \label{fig:HPA1032_opt_result}
\end{figure}

We evaluate the optimization performance of the proposed FMQA methods, LHS-FMQA and Sobol'-FMQA, on HPA103-1 and HPA103-2, which are single-objective unconstrained optimization benchmarks with implicit constraints handled internally by the simulator. The proposed methods are compared with Conv-FMQA, in which the initial training data are generated by uniform random sampling, as well as with Random Search and classical optimization methods, namely GP-BO~\cite{Shahriari2016BOReview,Jones1998EGO} and NSGA-II~\cite{Deb2002NSGAII}, under the same number of BB function evaluations. For HPA103-1 and HPA103-2, optimization is performed using ten different initial training datasets, and the average over ten trials of the best value found at each function evaluation is shown in Figs.~\ref{fig:HPA1031_opt_result} and \ref{fig:HPA1032_opt_result}. The horizontal axis represents the number of function evaluations, and the vertical axis represents the best value found. A larger value on the vertical axis as the evaluations proceed indicates that a higher cruising speed has been obtained.

The quantitative results are summarized in Table~\ref{tab:final_hpa103}. 
All reported values are cruising speeds in m/s. The values in parentheses in the ``Final best'' columns denote the difference from the average of Conv-FMQA.
\begin{table*}[t]
\centering
\caption{Initial and final best cruising speeds and gain for all methods in m/s. Values are mean $\pm$ standard deviation over ten trials. In the ``Final best'' columns, the values in parentheses denote the improvement over Conv-FMQA. Gain denotes the difference between the mean final and mean initial best cruising speeds.}
\label{tab:final_hpa103}
\renewcommand{\arraystretch}{1.0}
\scriptsize
\setlength{\tabcolsep}{4pt}
\begin{tabular*}{\textwidth}{@{\extracolsep{\fill}}lcccccc@{}}
\toprule
& \multicolumn{3}{c}{HPA103-1} & \multicolumn{3}{c}{HPA103-2} \\
\cmidrule(lr){2-4} \cmidrule(lr){5-7}
Method & Initial best & Final best & Gain & Initial best & Final best & Gain \\
\midrule
LHS-FMQA & $9.648 \pm 0.743$ & $10.698 \pm 0.359$ $(+0.135)$ & $1.050$ & $8.638 \pm 0.626$ & $10.328 \pm 0.343$ $(+0.333)$ & $1.690$ \\
Sobol'-FMQA & $9.703 \pm 0.373$ & $10.755 \pm 0.151$ $(+0.192)$ & $1.052$ & $8.929 \pm 0.468$ & $10.347 \pm 0.362$ $(+0.352)$ & $1.417$ \\
Conv-FMQA & $9.691 \pm 0.640$ & $10.563 \pm 0.274$ & $0.872$ & $8.824 \pm 0.421$ & $9.995 \pm 0.263$ & $1.171$ \\
GP-BO & $9.700 \pm 0.486$ & $10.589 \pm 0.486$ $(+0.027)$ & $0.889$ & $8.888 \pm 0.447$ & $10.262 \pm 0.176$ $(+0.267)$ & $1.374$ \\
NSGA-II & $9.691 \pm 0.607$ & $10.512 \pm 0.367$ $(-0.051)$ & $0.820$ & $8.984 \pm 0.502$ & $9.898 \pm 0.471$ $(-0.097)$ & $0.914$ \\
Random Search & $9.808 \pm 0.326$ & $10.113 \pm 0.222$ $(-0.450)$ & $0.305$ & $8.787 \pm 0.276$ & $9.395 \pm 0.342$ $(-0.600)$ & $0.609$ \\
\bottomrule
\end{tabular*}
\end{table*}

Figure~\ref{fig:HPA1031_opt_result} shows the optimization process for HPA103-1. Although the evaluation budget is fixed at 200 in this experiment, the behavior before reaching the final budget is also important because, when a target value is specified, reducing the number of BB function evaluations required to reach that value is a key objective in BBO. The proposed methods, LHS-FMQA and Sobol'-FMQA, increase rapidly after the initial training phase and outperform Conv-FMQA in the early stage of the optimization process. This indicates that the advantage of the proposed initial training data generation is particularly clear when the available evaluation budget is small. Conv-FMQA also improves as the evaluations proceed, but its increase is relatively gradual.

Table~\ref{tab:final_hpa103} summarizes the final values and the gains from the initial best values. On HPA103-1, LHS-FMQA and Sobol'-FMQA achieved higher mean final cruising speeds than all comparison methods, including GP-BO. 
However, the differences relative to GP-BO are smaller than the standard deviations reported in Table~\ref{tab:final_hpa103}, and we therefore do not claim a clear advantage over GP-BO on HPA103-1. 
The improvement over Conv-FMQA, which shares the same one-hot encoding and FM surrogate as the proposed methods, is the primary observation of this study.

Figure~\ref{fig:HPA1032_opt_result} shows the corresponding optimization process for HPA103-2, in which the number of design variables is approximately doubled. In this higher-dimensional case, the proposed methods show a rapid improvement immediately after the initial 32 evaluations, and their advantage over Conv-FMQA is most pronounced in the early-to-middle stage of the optimization process. After about 100 evaluations, the gap gradually decreases as Conv-FMQA also improves, but LHS-FMQA and Sobol'-FMQA still achieve higher final mean cruising speeds than Conv-FMQA and NSGA-II. Their final mean cruising speeds were comparable to those of GP-BO; the differences are less than 0.090 m/s and within the standard deviation. 
Comparing the results between HPA103-1 and HPA103-2, the relative advantage of the proposed methods over Conv-FMQA is more pronounced for the higher-dimensional problem: the improvement in final cruising speed grows from +0.135 m/s (LHS) and +0.192 m/s (Sobol') on HPA103-1 to +0.333 m/s (LHS) and +0.352 m/s (Sobol') on HPA103-2. 
Possible reasons for this dimension dependence are discussed in Sec.~\ref{Sec:Discussion}.

The initial-best and gain columns in Table~\ref{tab:final_hpa103} should be used to separate the quality of the initial samples from the improvement obtained during the subsequent optimization process. These columns show that the proposed methods do not simply start from uniformly better initial best values. On HPA103-1, the three FMQA variants begin from similar initial best cruising speeds around 9.65--9.70 m/s, but LHS-FMQA and Sobol'-FMQA achieve larger gains during the optimization process (1.050 and 1.052 m/s) than Conv-FMQA (0.872 m/s). 
On HPA103-2, this contrast is clearer. LHS-FMQA starts from a lower initial best cruising speed than Conv-FMQA (8.638 vs. 8.824 m/s), yet reaches a higher final best value because its gain during the optimization process is much larger (1.690 vs. 1.171 m/s).
Sobol'-FMQA, although starting from a slightly higher initial best value (8.929 m/s) than Conv-FMQA, likewise achieves a substantially larger gain (1.417 vs. 1.171 m/s). 
For both proposed methods, therefore, the improvement over Conv-FMQA is driven primarily by the optimization process after the initial sampling phase rather than by uniformly better initial samples.
The above observations suggest that ensuring marginal bit coverage in the initial training data improves the optimization performance of FMQA, particularly for higher-dimensional problems. We discuss the underlying mechanisms in Sec.~\ref{Sec:Discussion}.

\section{Discussion}
\label{Sec:Discussion}

\begin{figure}[b]
\centering

\begin{minipage}[b]{0.24\textwidth}
\centering
\includegraphics[width=\linewidth]{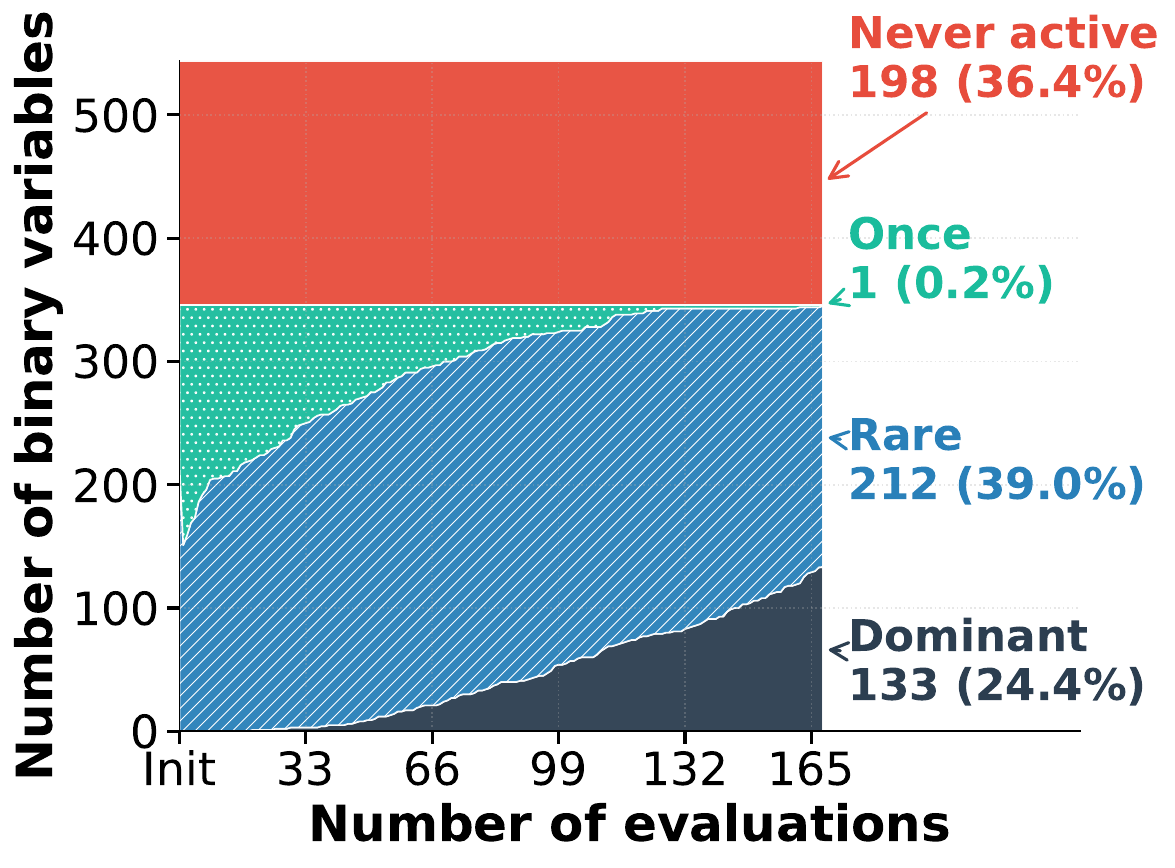}
\par\small (a) Conv-FMQA
\end{minipage}

\vspace{3mm}

\begin{minipage}[b]{0.48\linewidth}
\centering
\includegraphics[width=\linewidth]{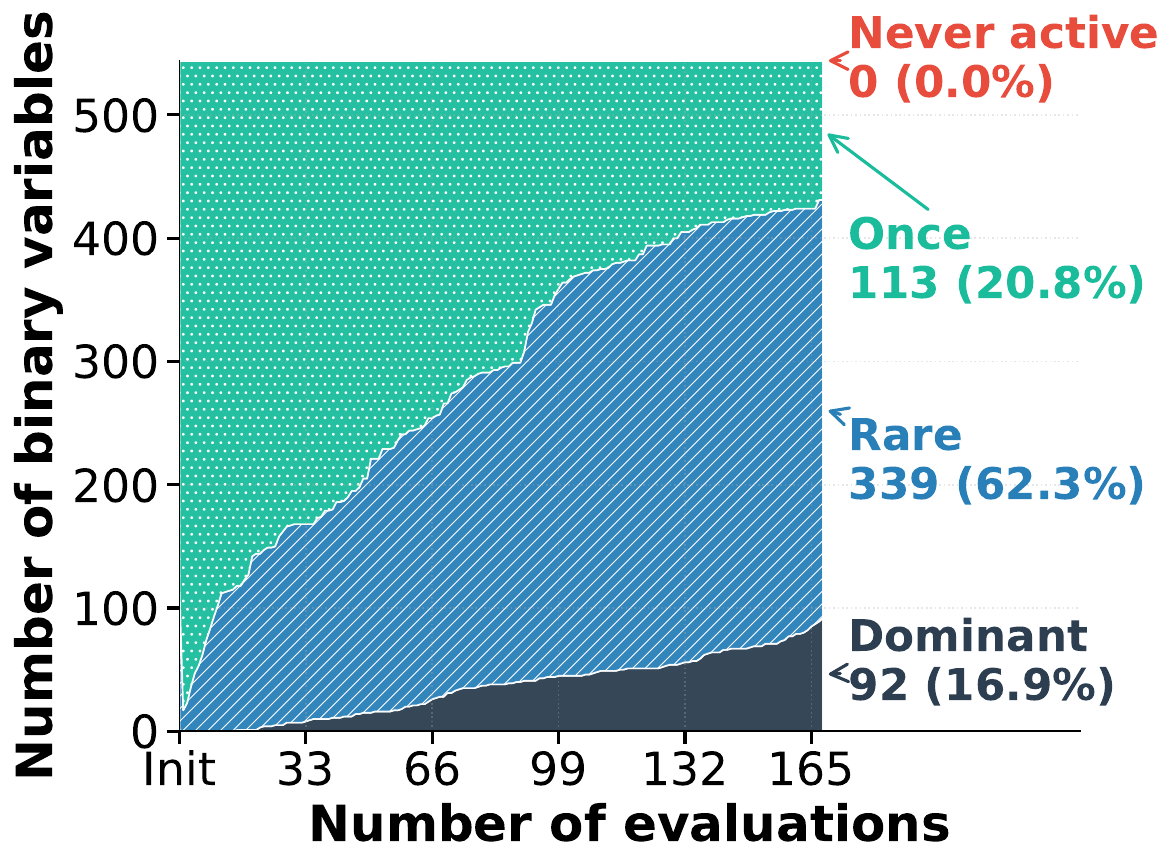}
\par\small (b) LHS-FMQA
\end{minipage}\hfill
\begin{minipage}[b]{0.48\linewidth}
\centering
\includegraphics[width=\linewidth]{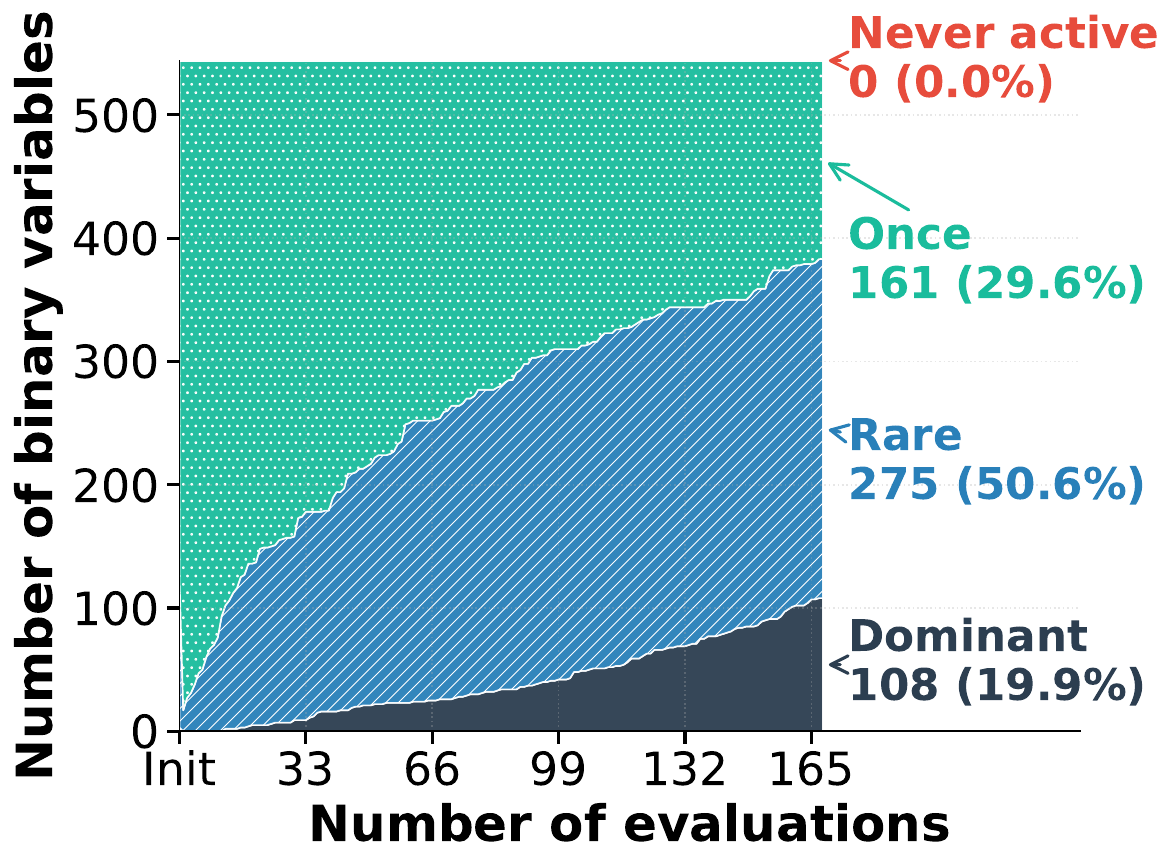}
\par\small (c) Sobol'-FMQA
\end{minipage}

\caption{
Comparison of the number of times each bit takes the value one in the bit arrays obtained by one-hot encoding for the datasets generated by (a) Conv-FMQA, (b) LHS-FMQA, and (c) Sobol'-FMQA on HPA103-1.
The horizontal axis represents the number of function evaluations, and the vertical axis represents the distribution of the number of times each bit in the dataset takes the value one at each function evaluation.
Red indicates bits that never take the value one, green indicates bits that take the value one only once, blue indicates bits that take the value one two to nine times, and black indicates bits that take the value one ten or more times.
}
\label{fig:activation_count_comparison_hpa1031}
\end{figure}

\begin{figure}[t]
\centering

\begin{minipage}[b]{0.24\textwidth}
\centering
\includegraphics[width=\linewidth]{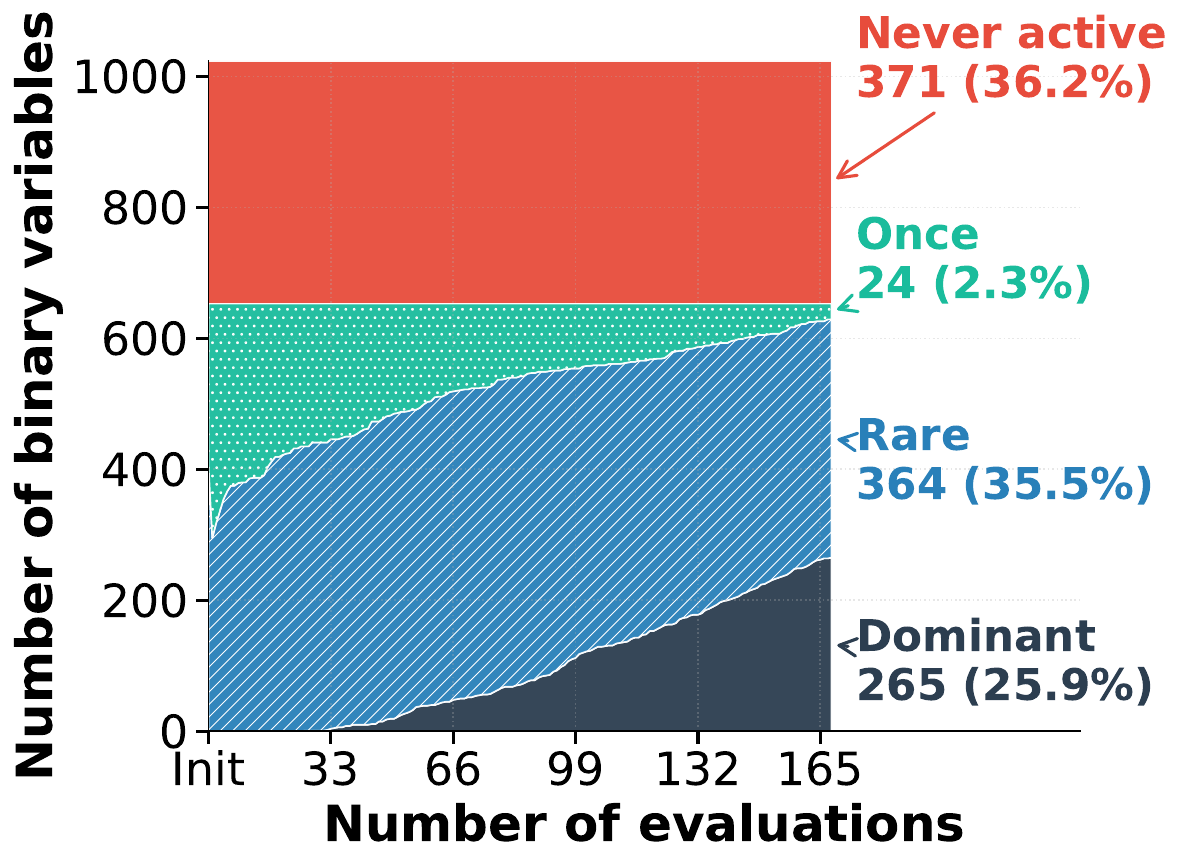}
\par\small (a) Conv-FMQA
\end{minipage}

\vspace{3mm}

\begin{minipage}[b]{0.48\linewidth}
\centering
\includegraphics[width=\linewidth]{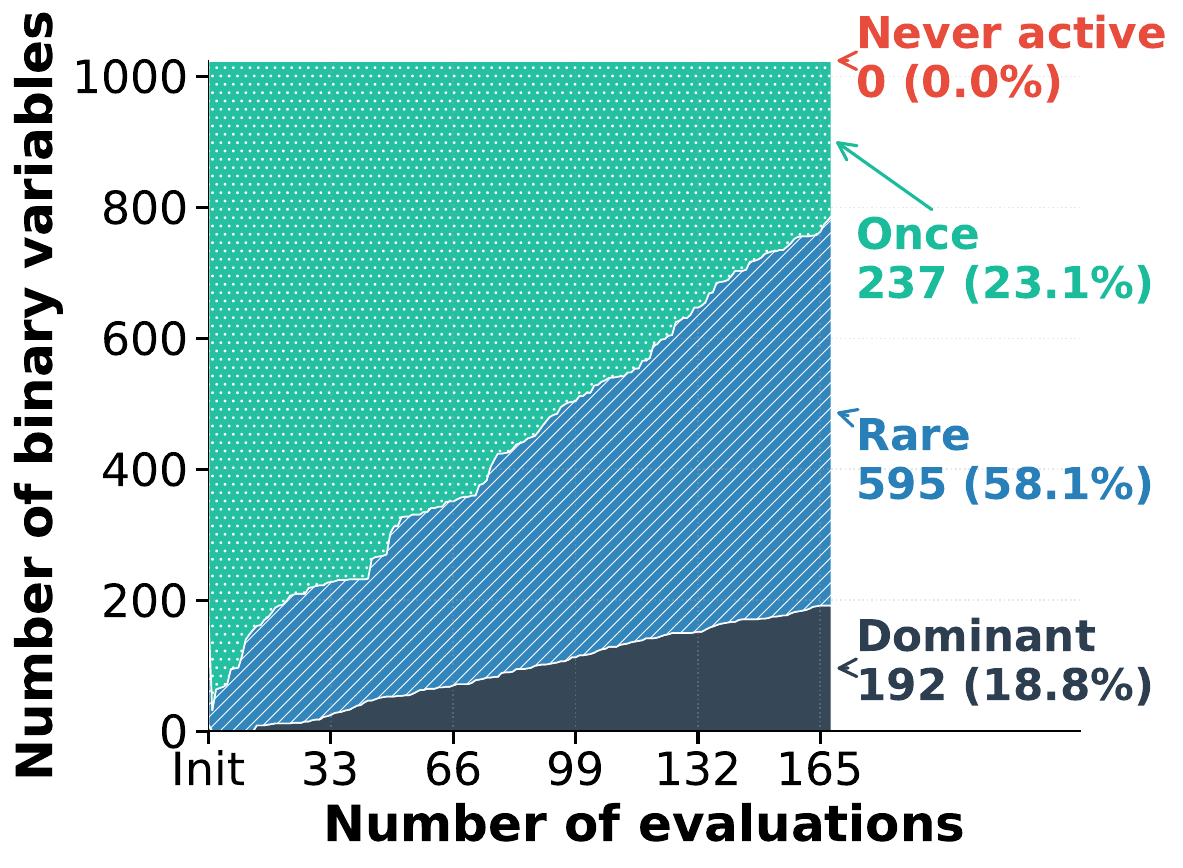}
\par\small (b) LHS-FMQA
\end{minipage}\hfill
\begin{minipage}[b]{0.48\linewidth}
\centering
\includegraphics[width=\linewidth]{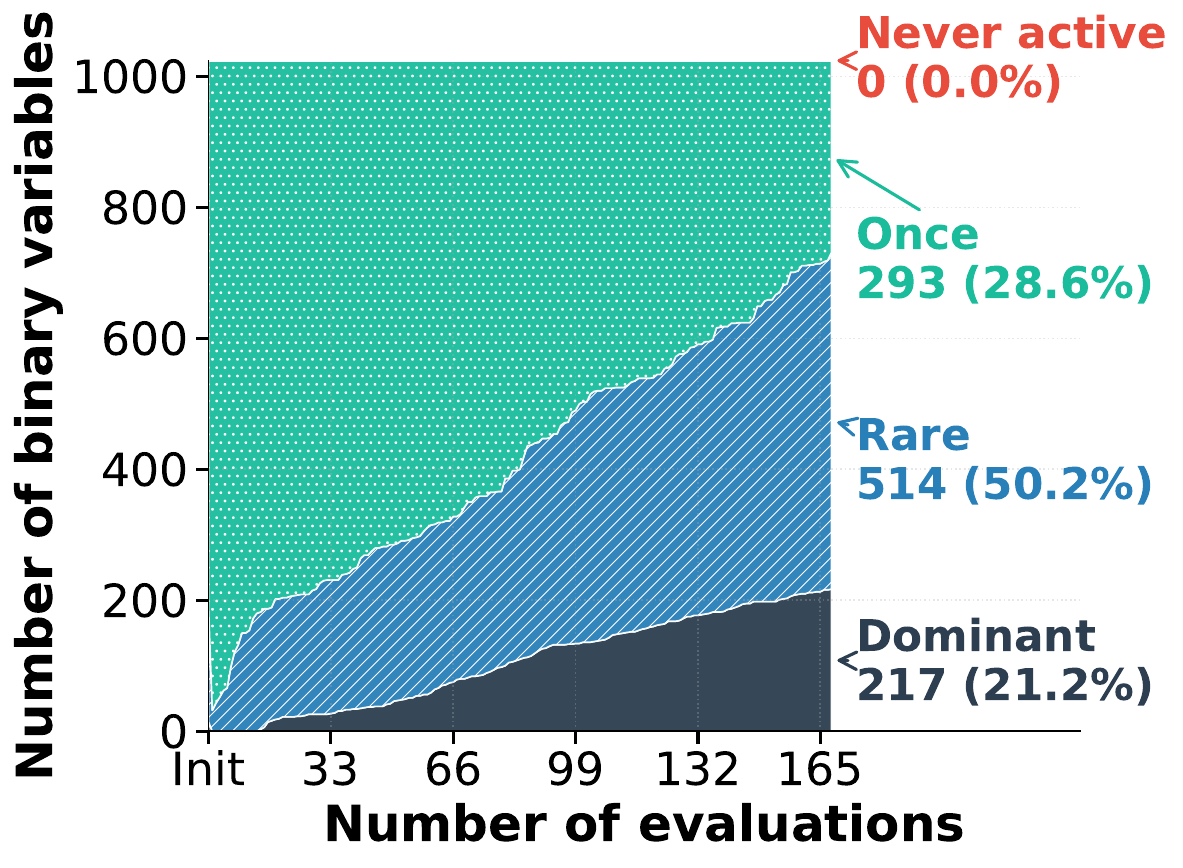}
\par\small (c) Sobol'-FMQA
\end{minipage}

\caption{
Comparison of the number of times each bit takes the value one in the bit arrays obtained by one-hot encoding for the datasets generated by (a) Conv-FMQA, (b) LHS-FMQA, and (c) Sobol'-FMQA on HPA103-2.
The horizontal axis represents the number of function evaluations, and the vertical axis represents the distribution of the number of times each bit in the dataset takes the value one at each function evaluation.
Red indicates bits that never take the value one, green indicates bits that take the value one only once, blue indicates bits that take the value one two to nine times, and black indicates bits that take the value one ten or more times.
}
\label{fig:activation_count_comparison_hpa1032}
\end{figure}

For HPA103-1 and HPA103-2, LHS-FMQA and Sobol'-FMQA achieved higher final mean cruising speeds than Conv-FMQA. In this section, we focus on how the initial training data generation method affects the distribution of bit occurrences in the dataset and discuss why the proposed methods can improve FMQA performance. Figures~\ref{fig:activation_count_comparison_hpa1031} and \ref{fig:activation_count_comparison_hpa1032} show the distributions of the number of times each bit in the dataset takes the value one as the number of function evaluations increases. By comparing these distributions, we examine how bit bias in the initial training data affects FM training and solution search by the Ising machine.


Figures~\ref{fig:activation_count_comparison_hpa1031} and \ref{fig:activation_count_comparison_hpa1032} show that, in LHS-FMQA and Sobol'-FMQA, all bits take the value one at least once. In contrast, in Conv-FMQA, bits that never take the value one remain even after the number of function evaluations increases. This tendency is consistent with the baseline estimate for uniform random sampling derived in the proposed-method section. As shown in Sec.~\ref{subsec:integer-to-binary-conversion}, when the initial training data are generated uniformly at random with $N_0=32$, about 36.4\% of the binary variables are expected never to take the value one. Figures~\ref{fig:activation_count_comparison_hpa1031} and \ref{fig:activation_count_comparison_hpa1032} show that such binary variables indeed remain in Conv-FMQA, and this tendency becomes more pronounced as the number of design variables increases, as is illustrated by HPA103-2. Consistently, Figs.~\ref{fig:HPA1031_opt_result} and \ref{fig:HPA1032_opt_result} and Table~\ref{tab:final_hpa103} show that LHS-FMQA and Sobol'-FMQA attain higher final objective values than Conv-FMQA.

These observations are consistent with the following mechanism. As discussed in Sec.~\ref{Subsec:variety_of_bits}, FM parameters corresponding to bits that never take the value one do not receive direct updates from dataset-derived gradients. In Conv-FMQA, FM parameter updates may therefore be biased toward bits that take the value one in the initial training data, and the subsequent Ising-machine search may also remain concentrated around combinations of such bits. By contrast, in LHS-FMQA and Sobol'-FMQA, the initial training data are generated so that every binary variable obtained by one-hot encoding takes the value one at least once. This suppresses the occurrence of FM parameters that are not informed by dataset-derived gradients and may mitigate bias in both parameter updates and solution search. In this sense, the present results support the hypothesis that improving the marginal coverage of the binary variables obtained by one-hot encoding in the initial training data improves FMQA performance.

At the same time, the current analysis has clear limitations. We did not observe an appreciable performance difference between LHS-FMQA and Sobol'-FMQA.
One possible interpretation is that, once complete marginal bit coverage is achieved, higher-order space-filling properties in the original continuous design space contribute little to the FMQA performance under the present problem and budget. 
At the same time, both LHS and Sobol' provide space-filling structure as well as marginal coverage, so the present comparison alone cannot determine whether marginal coverage is the dominant factor or whether the contributions of these two properties happen to be of similar magnitude. 
Resolving this would require an initialization scheme that achieves marginal coverage without space-filling structure, which we leave for future work.
However, the proposed initialization methods simultaneously achieve two distinct properties: (i) complete marginal coverage of the binary variables obtained by one-hot encoding, which is the focus of this study, and (ii) a space-filling distribution in the original continuous design space. Our experiments do not fully isolate the effect of (i) from that of (ii). Moreover, even in LHS-FMQA and Sobol'-FMQA, a fraction of binary variables (dominant binary variables, 17--21\% of all binary variables) take the value one ten or more times by the end of the optimization process, indicating that the candidate search later becomes concentrated on a subset of bit combinations.

These limitations lead to several open questions. It remains unclear how much of the observed performance gain is attributable specifically to marginal bit coverage rather than to the space-filling distribution of the initial samples in the original continuous design space. It is also unclear whether the concentration of bit usage observed in the later stage of the optimization process reflects desirable convergence toward good solutions or undesirable local-search bias. Addressing these questions requires additional controlled comparisons, for example against an initialization scheme that guarantees complete marginal coverage without introducing the same space-filling property. Such analyses are left for future work.

    \section{Conclusion and Future Perspectives}
\label{Sec:Conclusion}
In this study, we proposed initial training data generation methods for FMQA with one-hot encoding. 
When one-hot encoding is used together with uniform random initial sampling, some binary variables may never take the value one in the initial training data, and the corresponding FM parameters do not receive direct gradient updates from the observed BB responses. 
To address this issue, we focused on complete marginal bit coverage, namely, ensuring that every binary variable obtained by one-hot encoding takes the value one at least once in the initial training dataset. 
To realize this condition, we used LHS and the Sobol' sequence as practical sampling procedures and incorporated them into the FMQA framework.

Numerical experiments on the human-powered aircraft wing-shape optimization benchmark showed that LHS-FMQA and Sobol'-FMQA achieved numerically higher final mean cruising speeds than Conv-FMQA, particularly on the higher-dimensional HPA103-2.
The initial-best and gain values indicate that this difference was not simply due to uniformly better initial best values, but reflects the optimization process after the initial sampling phase. 
The bit-usage analysis also confirmed that the proposed methods achieved complete marginal bit coverage, whereas about 36\% of the binary variables in Conv-FMQA never took the value one even after the full evaluation budget was exhausted, consistent with the $(1-1/M)^{N_0}$ estimate. 
In the present benchmark and experimental setting, no appreciable performance difference was observed between LHS-FMQA and Sobol'-FMQA.

These results are consistent with the hypothesis that improving marginal bit coverage reduces the number of FM parameters that are not informed by dataset-derived gradients and thereby mitigates bias in subsequent QUBO-based search. 
Under this interpretation, the QUBO matrix passed to the Ising machine carries observation-derived information along all design dimensions, rather than being effectively uninformed along the missing-bit directions. 
This effect is expected to be more pronounced in higher-dimensional problems, where the absolute number of never-active bits in Conv-FMQA grows with the search dimension.

Several limitations should be acknowledged. 
First, the proposed sampling procedures simultaneously achieve complete marginal bit coverage and a space-filling distribution in the original continuous design space, and the present experiments do not fully separate the contribution of these two properties. 
The above interpretation should therefore be regarded as a hypothesis rather than a complete causal explanation. 
Second, the proposed methods guarantee only marginal coverage and do not guarantee coverage of pairwise bit combinations, which directly informs the FM interaction parameters. 
Third, our guarantee assumes $N_0 = M$ (with the additional condition $M = 2^p$ for the Sobol' construction), which means that the initial evaluation cost scales with $M$; when the BB function is extremely expensive and only a small $N_0$ is affordable, partial coverage strategies will be required. 
Finally, it remains unclear whether the concentration of bit usage in the later stage of the optimization process reflects desirable convergence or undesirable local-search bias.

Several directions for future work follow from these limitations. 
A coverage-only initialization that achieves complete marginal bit coverage without space-filling structure in the continuous design space would isolate the effect of marginal coverage and directly test our hypothesis. 
Initialization schemes that guarantee pairwise bit coverage, as well as adaptive schemes that prioritize the most influential variables when $N_0 < M$, are also worth investigating. 
Broader benchmark studies that relate the effectiveness of such designs to the structure of the BB function (e.g., multimodality, separability) and statistical evaluation across more trials are required to establish the generality of our findings. 
In the longer term, these analyses could lead to practical design rules for choosing initial training data in FMQA and related surrogate-model-based BBO methods.

    \begin{acknowledgments}
        
        \section*{Acknowledgments}        
        This work was partially supported by the Japan Society for the Promotion of Science (JSPS) KAKENHI (Grant Number JP23H05447), the Council for Science, Technology, and Innovation (CSTI) through the Cross-ministerial Strategic Innovation Promotion Program (SIP), ``Promoting the application of advanced quantum technology platforms to social issues'' (Funding agency: QST), Japan Science and Technology Agency (JST) (Grant Number JPMJPF2221).
        S.~Tanaka wishes to express their gratitude to the World Premier International Research Center Initiative (WPI), MEXT, Japan, for their support of the Human Biology-Microbiome-Quantum Research Center (Bio2Q).
    \end{acknowledgments}

    \bibliographystyle{jpsj}
    \bibliography{reference}
\end{document}